\documentclass[10pt,twocolumn,letterpaper]{article}

\usepackage{iccv}    

\iccvfinalcopy

\usepackage{graphicx}
\usepackage{amsmath}
\usepackage{amssymb}
\usepackage{booktabs}

\usepackage{xcolor} 
\usepackage{times}
\usepackage{epsfig}
\usepackage{graphicx}
\usepackage{amsmath}
\usepackage{amssymb}
\usepackage[bb=dsserif]{mathalpha}
\usepackage{bm}

\usepackage{booktabs,colortbl,tabularx}
\usepackage{pifont}%

\usepackage{mathtools}
\usepackage{commath}
\usepackage{algorithm}
\usepackage[noend]{algpseudocode}

\usepackage{multirow}
\usepackage{comment} 
\usepackage{enumitem}

\usepackage[accsupp]{axessibility}

\definecolor{Gray}{gray}{0.9}

\newcommand{\cmark}{\ding{51}}%
\newcommand{\xmark}{\ding{55}}%

\definecolor{battleshipgrey}{rgb}{0.52, 0.52, 0.51}

\usepackage[pagebackref,breaklinks,colorlinks]{hyperref}

\usepackage[capitalize]{cleveref}
\crefname{section}{Sec.}{Secs.}
\Crefname{section}{Section}{Sections}
\Crefname{table}{Table}{Tables}
\crefname{table}{Tab.}{Tabs.}

\ificcvfinal\fi

\begin{document}

\title{Class-Incremental Grouping Network for \\ Continual Audio-Visual Learning}

\author{%
  Shentong Mo$^\dagger$\\
  Carnegie Mellon University
  \and
  Weiguo Pian$^\dagger$\\
  University of Texas at Dallas
  \and
  Yapeng Tian\thanks{Corresponding author, $^\dagger$Equal contribution.}\\
  University of Texas at Dallas
}

\maketitle

\begin{abstract}

Continual learning is a challenging problem in which models need to be trained on non-stationary data across sequential tasks for class-incremental learning. 
While previous methods have focused on using either regularization or rehearsal-based frameworks to alleviate catastrophic forgetting in image classification, they are limited to a single modality and cannot learn compact class-aware cross-modal representations for continual audio-visual learning.
To address this gap, we propose a novel class-incremental grouping network (CIGN) that can learn category-wise semantic features to achieve continual audio-visual learning. 
Our CIGN leverages learnable audio-visual class tokens and audio-visual grouping to continually aggregate class-aware features. 
Additionally, it utilizes class tokens distillation and continual grouping to prevent forgetting parameters learned from previous tasks, thereby improving the model's ability to capture discriminative audio-visual categories.
We conduct extensive experiments on VGGSound-Instruments, VGGSound-100, and VGG-Sound Sources benchmarks.
Our experimental results demonstrate that the CIGN achieves state-of-the-art audio-visual class-incremental learning performance.
Code is available at \url{https://github.com/stoneMo/CIGN}.

\end{abstract}


\vspace{-0.5em}
\section{Introduction}
\vspace{-0.5em}

The strong correspondence between audio signals and visual objects in the world enables humans to perceive the source of a sound, such as a meowing cat. 
This perception intelligence has motivated researchers to explore audio-visual joint learning for various tasks, such as audio-visual event classification~\cite{tian2018ave}, sound source separation~\cite{zhao2018the,Gao2018learning,tian2021cyclic}, and visual sound localization~\cite{Senocak2018learning,chen2021localizing,mo2022EZVSL}.
In this work, we focus on the problem of classifying sound sources from both video frames and audio in a continual learning~\cite{Kirkpatrick2017overcoming,McCloskey1989CatastrophicII} setting, \textit{i.e.}, train audio-visual learning models on non-stationary audio-visual pairs, enabling the model to classify sound sources in videos incrementally.

\begin{figure}
\centering
\includegraphics[width=1.0\linewidth]{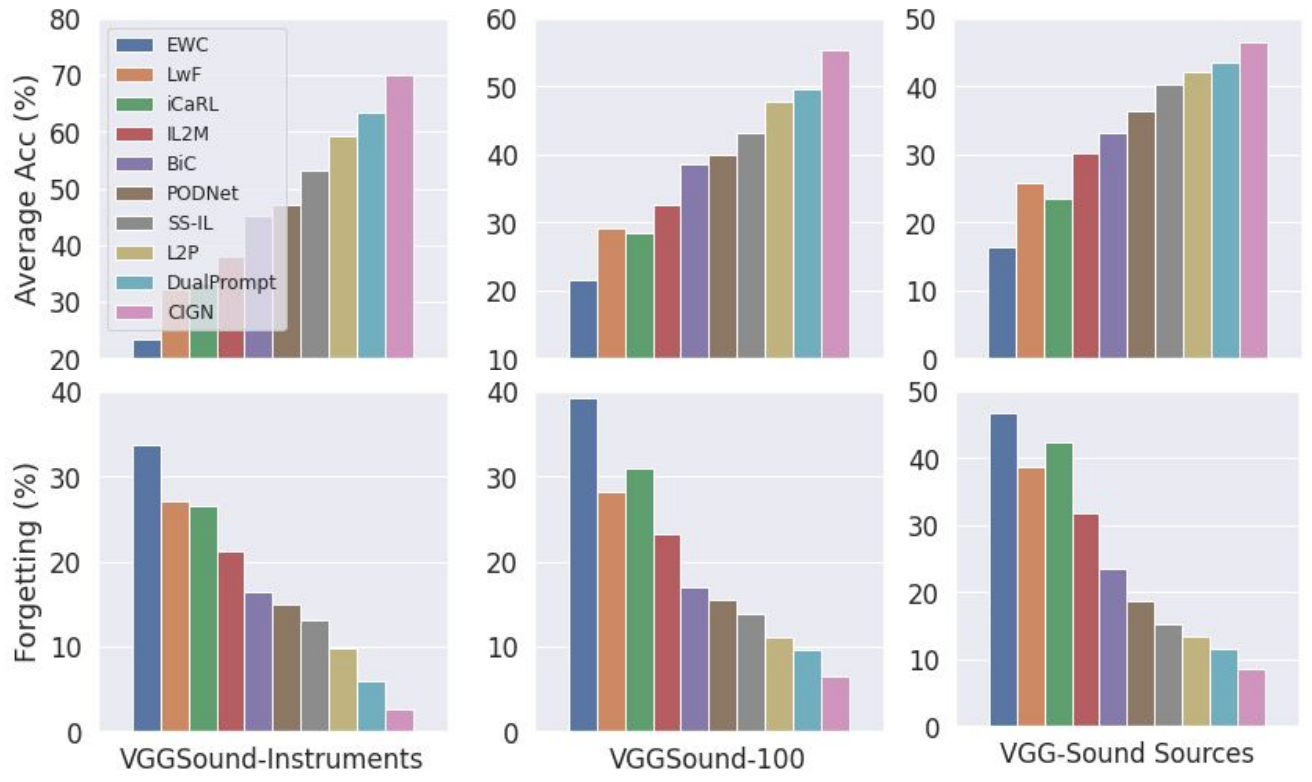}
\caption{Comparison of our CIGN with state-of-the-art approaches on Average Accuracy (Top Row, higher is better) and Forgetting (Bottom Row, lower is better) for continual audio-visual learning on VGGSound-Instruments~\cite{hu2022mix}, VGGSound-100~\cite{chen2020vggsound}, and VGG-Sound Sources~\cite{chen2020vggsound} benchmarks. }
\label{fig: title_img}
\end{figure}

Continual learning is a recently popular and challenging problem that aims to train models on non-stationary data given sequential tasks for class-incremental learning.
Previous methods~\cite{Kirkpatrick2017overcoming,Rebuffi2017icarl,li2018learning,Castro2018eeil,wu2019large,Hou2019learning,prabhu2020greedy} mainly used either regularization or rehearsal-based approaches to alleviate catastrophic forgetting between old and new classes for image classification. 
A typical regularization-based work, 
EWC~\cite{Kirkpatrick2017overcoming} addressed catastrophic forgetting in neural networks by training sequential models to protect the weights crucial for previous tasks.
Similarly, LwF~\cite{li2018learning} used only new task data for training while preserving the original capabilities.
To improve the performance under challenging datasets, BiC~\cite{wu2019large} adopted two bias
parameters in a linear model to address
the bias to new classes.
SS-IL~\cite{Ahn2021ssil} combined separated softmax with task-wise knowledge distillation to solve this bias. 
However, those baselines are based on a single input modality and perform worse for continual audio-visual learning. 
In this work, we address a new multi-modal continual problem by extracting disentangled and compact representations with learnable audio-visual class-incremental tokens as guidance for continual learning.

Since prompts can learn task-specific knowledge~\cite{Pfeiffer2020AdapterFusionNT,lester2021power,Li2021PrefixTuningOC,hu2022lora} for transfer learning, recent researchers have tried to adapt diverse prompts-based pipelines to resolve continual learning challenges.
The basic idea of prompting is to design a function to alter the input text for pre-trained large language models such that the model can generate more informative features about the new task.
For example, L2P~\cite{wang2022l2p} leveraged a prompt pool memory space to instruct the model prediction and explicitly maintain model plasticity for task-invariant and task-specific knowledge.
More recently, DualPrompt~\cite{wang2022dualprompt} proposed complementary prompts to instruct the pre-trained backbone for learning tasks arriving sequentially.
Despite their promising results, they can only deal with one modality, and the design of audio-visual prompts requires choreographed heuristics.
When we apply their prompting to our audio-visual settings, they cannot learn compact class-aware cross-modal representations for class-incremental learning.

The main challenge is that sounds are naturally mixed in the audio space such that the global audio representation extracted from the sound is easy to be catastrophically forgotten by the
cross-modal model.
This inspires us to disentangle the individual semantics for old and current audio-visual pairs to guide continual audio-visual learning. 
To address the problem, our key idea is to disentangle individual audio-visual representations from sequential tasks using audio-visual continual grouping for class-incremental learning, which differs from existing regularization-based and rehearsal-based methods. 
During training, we aim to learn audio-visual class tokens to continually aggregate category-aware source features from the sound and the image, where separated high-level semantics for sequential audio-visual tasks are learned.

To this end, we propose a novel class-incremental grouping network, namely CIGN, that can directly learn category-wise semantic features to achieve continual audio-visual learning.
Specifically, our CIGN leverages learnable audio-visual class tokens and audio-visual grouping to continually aggregate class-aware features.
Furthermore, it leverages audio-visual class tokens distillation and continual grouping to alleviate forgetting parameters learned from previous tasks for capturing discriminative audio-visual categories.

Empirical experiments on VGGSound-Instruments, VGGSound-100, and VGG-Sound Sources benchmarks comprehensively demonstrate the state-of-the-art performance against previous continual learning baselines, as shown in Figure~\ref{fig: title_img}. 
In addition, qualitative visualizations of learned audio-visual embeddings vividly showcase the effectiveness of our CIGN in aggregating class-aware features to avoid cross-modal catastrophic forgetting. 
Extensive ablation studies also validate the importance of class-token distillation and continual grouping in learning compact representations for class-incremental learning.

Our contributions can be summarized as follows:
\begin{itemize}
    \item We present a novel class-incremental grouping network, namely CIGN, that can directly learn category-wise semantic features to achieve continual audio-visual learning.
    \item We introduce learnable audio-visual class tokens distillation and continual grouping to continually aggregate class-aware features for alleviating cross-modal catastrophic forgetting.
    \item Extensive experiments can demonstrate the state-of-the-art superiority of our CIGN over previous baselines on audio-visual class-incremental scenarios.
\end{itemize}

\section{Related Work}

\noindent\textbf{Audio-Visual Learning.}
Audio-visual learning has been explored in many previous methods~\cite{aytar2016soundnet,owens2016ambient,Arandjelovic2017look,korbar2018cooperative,Senocak2018learning,zhao2018the,zhao2019the,Gan2020music,su2020audeo,su2021does,Morgado2020learning,Morgado2021robust,Morgado2021audio,mo2023diffava,mo2023unified,li2022learning} to capture audio-visual alignment from those two different modalities in videos. A comprehensive survey on audio-visual learning can be found in \cite{wei2022learning}.
Given video sequences with audio and frames, the objective is to push away embeddings from non-matching audio-visual pairs while closing audio and visual representations from the matching pair.
Such cross-modal alignment is helpful for several tasks, such as audio spatialization~\cite{Morgado2018selfsupervised,gao20192.5D,Chen2020SoundSpacesAN,Morgado2020learning}, audio/speech separation~\cite{Gan2020music,Gao2018learning,Gao2019co,zhao2018the,zhao2019the,gao2020listen,tian2021cyclic,gao2021visualvoice}, visual sound source localization~\cite{Senocak2018learning,Rouditchenko2019SelfsupervisedAC,hu2019deep,Afouras2020selfsupervised,qian2020multiple,chen2021localizing,mo2022EZVSL,mo2022SLAVC,mo2023audiovisual,mo2023avsam}, and audio-visual event parsing~\cite{tian2018ave, tian2020avvp,wu2021explore,lin2021exploring,mo2022multimodal}.
In this work, our main focus is to learn compact audio-visual representations on non-stationary audio-visual pairs given sequential tasks for class-incremental learning, which is more challenging than the abovementioned tasks on independent and identically distributed audio-visual data.

\begin{figure*}[t]
\centering
\includegraphics[width=0.92\linewidth]{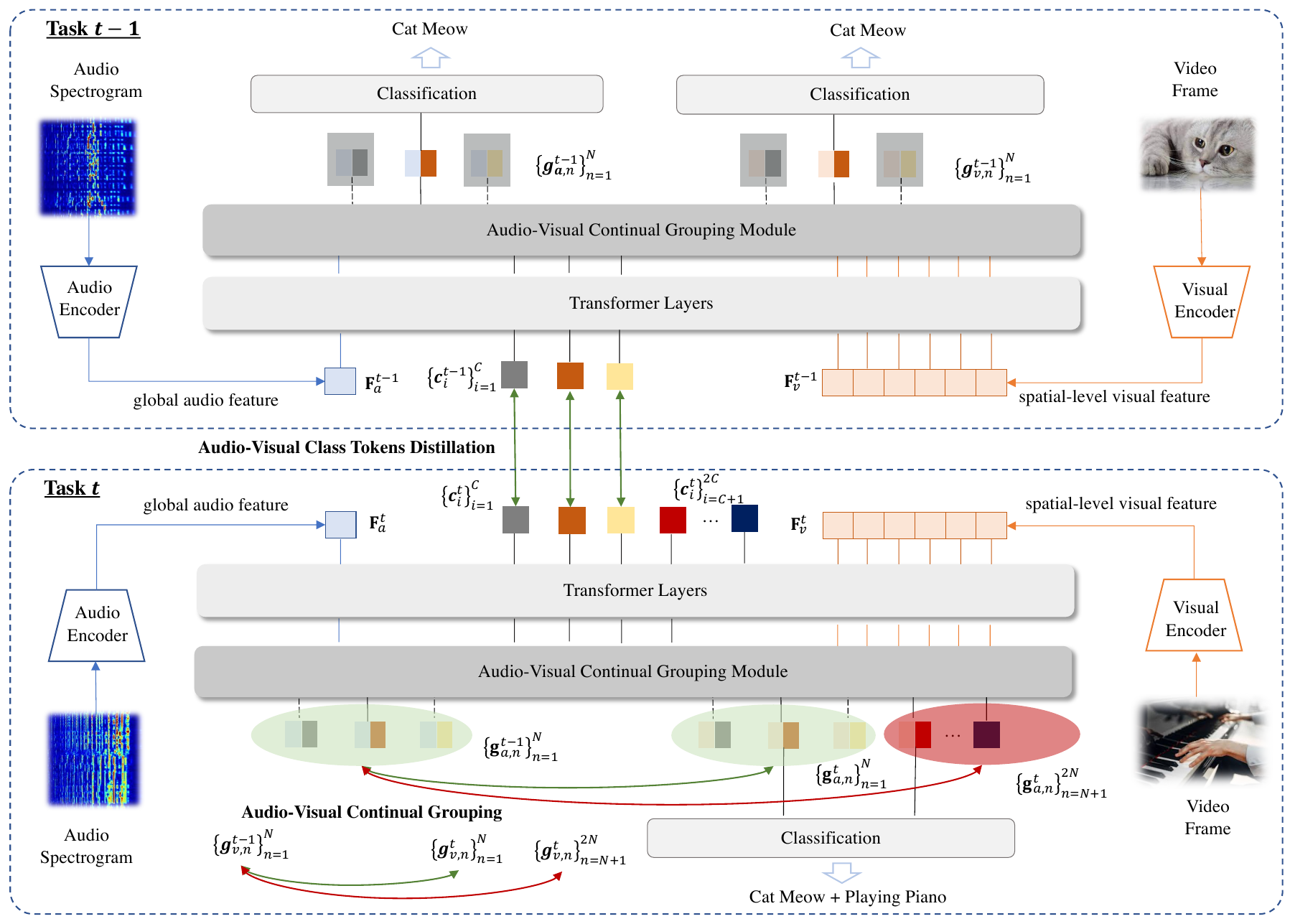}
\caption{Illustration of the proposed Class-Incremental Grouping Network (CIGN).
The Audio-Visual Continual Grouping module takes as audio features $\mathbf{F}_a^t=\mathbf{f}_a^t$ of the spectrogram, visual features $\mathbf{F}_v^t=\{\mathbf{f}^t_{v,p}\}_{p=1}^P$ of the video frame from each encoder and learnable audio-visual class tokens $\{\mathbf{c}_i^t\}^{2C}_{i=1}$ for $2C$ classes in the semantic space to generate disentangled class-aware audio-visual representations $\{\mathbf{g}^t_{a,n}\}^{2N}_{n=1}, \{\mathbf{g}^t_{v,n}\}^{2N}_{n=1}$ for $2N$ sources at task $t$.
Note that $2N$ source embeddings are chosen from $2C$ categories according to the ground-truth class.
The Audio-Visual Class Tokens Distillation module constrains the distribution of old class tokens $\{\mathbf{c}_i^t\}^{C}_{i=1}$ at task $t$ same as those $\{\mathbf{c}_i^{t-1}\}^{C}_{i=1}$ at task $t-1$.
Meanwhile, class-aware audio-visual features $\{\mathbf{g}^{t-1}_{a,n}\}^{N}_{n=1}, \{\mathbf{g}^{t-1}_{v,n}\}^{N}_{n=1}$ at task $t-1$ are pulled close (Green arrow) to the associated class-aware features $\{\mathbf{g}^{t}_{a,n}\}^{N}_{n=1}, \{\mathbf{g}^{t}_{v,n}\}^{N}_{n=1}$ and away (Red arrow) from those features with different classes $\{\mathbf{g}^{t}_{a,n}\}^{2N}_{n=N+1}, \{\mathbf{g}^{t}_{v,n}\}^{2N}_{n=N+1}$ at task $t$. 
Finally, the classification layer composed of an FC layer and a sigmoid function is separately used to predict audio and video classes, where the product of FC output logits is utilized to generate audio-visual categories.
}
\label{fig: main_img}
\end{figure*}

\vspace{2mm}
\noindent\textbf{Continual Learning.}
Continual learning aims to train a single model on non-stationary data distributions where distinct classification tasks are given sequentially.
Early works~\cite{Kirkpatrick2017overcoming,li2018learning,Aljundi2018memory} mainly applied regularization on the learning rate on crucial parameters for old tasks to preserve the model capability.
In recent years, diverse rehearsal-based pipelines~\cite{Rebuffi2017icarl,Castro2018eeil,Belouadah2019il2m,Hou2019learning,chaudhry2019er,wu2019large,prabhu2020greedy,Buzzega2020dark,cha2021co2l,Ahn2021ssil} have been proposed to resolve the catastrophic forgetting problem in challenging settings.
A dual memory network was proposed in IL2M~\cite{Belouadah2019il2m} to use both a bounded memory of the past images and a second memory for past
class statistics obtained from initial training. 
By carefully balancing the compromise between old and new classes, PODNet~\cite{Douillard2020podnet} introduced a spatial-based distillation loss to optimize the model with multiple proxy embeddings for each category.
Unlike them, we address the cross-modal catastrophic forgetting problem in audio-visual settings.  
Instead, we leverage the category-aware representations of individual audio-visual pairs to predict the corresponding category for sequential classification tasks, where learnable audio-visual class tokens are utilized as the desirable guidance for class-incremental audio-visual learning.

Due to the promising performance of prompts in transfer learning, recent works~\cite{wang2022l2p,wang2022dualprompt} also have explored different frameworks to alleviate catastrophic forgetting in class-incremental visual learning.
A prompt pool memory-based framework was designed in L2P~\cite{wang2022l2p} to prompt a pre-trained vision transformer~\cite{dosovitskiy2020image} for learning sequential classification tasks dynamically. 
Inspired by the Complementary Learning Systems~\cite{McClelland1995WhyTA,Kumaran2016WhatLS} theory, DualPrompt~\cite{wang2022dualprompt} combined General-Prompt and Expert-Prompt for learning task-invariant and task-specific knowledge.
While those prompts-based approaches achieve promising performance in continual visual learning, they can only handle the texts targeted for images, and audio-visual prompts need to be well-designed.
Applying their prompting to our audio-visual settings fails to learn compact class-aware cross-modal representations for audio-visual class-incremental learning~\cite{pian2023audiovisual}.
In contrast, we develop a fully novel framework to aggregate compact category-wise audio and visual source representations with explicit learnable source class tokens. 
To the best of our knowledge, we are the first to leverage an explicit grouping mechanism for continual audio-visual learning.
Our experiments in Section~\ref{sec:exp} also validate the superiority of CIGN in all benchmarks for class-incremental audio-visual settings.

\section{Method}

Given an image and a spectrogram of audio, our target is to classify the visual sound source sequentially for continual audio-visual learning. 
We propose a novel class-incremental grouping network, named CIGN, for disentangling individual semantics from the audio and image, which mainly consists of two modules, Audio-Visual Class Tokens Distillation in Section~\ref{sec:avctd} and Audio-Visual Continual Grouping in Section~\ref{sec:avcg}.

\subsection{Preliminaries}

In this section, we first describe the problem setup and notations and then revisit the audio-visual class-incremental classification for continual learning.

\noindent\textbf{Problem Setup and Notations.}
Given a spectrogram and an image, our goal is to continually classify non-stationary audio-visual pairs given sequential tasks for class-incremental learning.
We have an audio-visual label for a video with $C$ audio-visual event categories, denoted as $\{y_i\}_{i=1}^C$ with $y_i$ for the ground-truth category entry $i$ as 1. 
During training, we have video-level annotations as supervision. 
Therefore, we can leverage the video-level label for the audio mixture spectrogram and image to perform continual audio-visual learning.

\noindent\textbf{Revisit Continual Visual Learning.}
Given a sequence of tasks $\mathcal{D} = \{\mathcal{D}_1, ..., \mathcal{D}_T\}$, where the $t$-th task $\mathcal{D}_t = \{(\mathbf{x}^t_i, y_i^t)\}_{i=1}^{n_t}$ contains tuples of 
the input sample $\mathbf{x}^t_i\in \mathcal{X}$ and its corresponding label $y_i^t\in \mathcal{Y}$.
The goal of continual visual learning is to train a single model with parameters $\theta$, \textit{i.e.}, $f_\theta: \mathcal{X}\rightarrow\mathcal{Y}$, which enables it to predict the label $y = f_\theta(\mathbf{x})\in \mathcal{Y}$ given an unseen test sample $\mathbf{x}$ from arbitrary tasks. 
Data from the previous tasks may not be seen anymore when training future tasks.
Different from traditional continual visual learning, we try to involve audio-visual pairs samples together in each sequential task.
Specifically, the $t$-th task $\mathcal{D}_t$ includes tuples of the input audio-visual pairs $\mathbf{a}^t_i\in \mathcal{A}, \mathbf{v}^t_i\in \mathcal{V}$ and its corresponding label $y_i^t\in \mathcal{Y}$.
Our target is to train a single multi-modal model with parameters $\theta$, \textit{i.e.}, 
$g_\theta: \mathcal{A}\times \mathcal{V}\rightarrow\mathcal{Y}$, such that we can predict the audio-visual event category $y = g_\theta(\mathbf{a},\mathbf{v})\in \mathcal{Y}$ given an unseen test audio-visual pair $\mathbf{a},\mathbf{v}$ from arbitrary tasks.
When training future tasks, audio-visual pairs from the previous tasks might not be seen anymore.

However, such a multi-modal continual learning setting will pose the main challenge for class-incremental audio-visual classification.
The global audio representation extracted from the sound is catastrophically forgotten by the cross-modal model, and thus they can not associate individual audio with the corresponding image in future audio-visual tasks.
To tackle the challenge, we are inspired by~\cite{xu2022groupvit} and propose a novel class-incremental grouping network, namely CIGN, for disentangling individual event semantics from the audio and image to achieve class-incremental audio-visual learning, as illustrated in Figure~\ref{fig: main_img}.

\subsection{Audio-Visual Class Tokens Distillation}\label{sec:avctd}
To explicitly learn disentangled semantics from incremental audio and images, we introduce a novel audio-visual class token distillation to constrain the distribution of previous class tokens $\{\mathbf{c}_i^t\}^{C}_{i=1}$ at task $t$ same as those $\{\mathbf{c}_i^{t-1}\}^{C}_{i=1}$ at task $t-1$.
The learnable audio-visual incremental class tokens $\{\mathbf{c}_i^t\}^{2C}_{i=1}$ at task $t$ further group semantic-aware information from audio-visual representations $\mathbf{f}^t_a, \{\mathbf{f}^t_{v,p}\}_{p=1}^P$, where $\mathbf{c}_i^t\in\mathbb{R}^{1\times D}$, 
$2C$ is the total number of old ($C$) and new ($C$) classes at task $t$, and $P$ denotes the number of total spatial locations in the feature map.

With the incremental categorical audio-visual tokens and raw representations, we first apply self-attention transformers $\phi^t_a(\cdot), \phi^t_v(\cdot)$ at task $t$ to aggregate global audio and spatial visual representations from the raw input and align the cross-modal features with the class token embeddings as:
\begin{equation}
\begin{aligned}
    & \hat{\mathbf{f}}^t_a, \{\hat{\mathbf{c}}^t_{a,i}\}_{i=1}^{2C} = \{\phi^t_a(\mathbf{x}^t_{a,j}, \mathbf{X}^t_a, \mathbf{X}^t_a)\}_{j=1}^{1+2C}, \\
    & \mathbf{X}^t_a = \{\mathbf{x}^t_{a,j}\}_{j=1}^{1+2C} = [\mathbf{f}^t_a; \{\mathbf{c}^t_i\}_{i=1}^{2C}]
\end{aligned}
\end{equation}
\begin{equation}
\begin{aligned}
    & \{\hat{\mathbf{f}}^t_{v,p}\}_{p=1}^P, \{\hat{\mathbf{c}}^t_{v,i}\}_{i=1}^{2C} = \{\phi^t_v(\mathbf{x}^t_{v,j}, \mathbf{X}^t_v, \mathbf{X}^t_v)\}_{j=1}^{P+2C}, \\
    & \mathbf{X}^v = \{\mathbf{x}^t_{v,j}\}_{j=1}^{P+2C} = [\{\mathbf{f}^t_{v,p}\}_{p=1}^P; \{\mathbf{c}^t_i\}_{i=1}^{2C}]
\end{aligned}
\end{equation}
where $\hat{\mathbf{f}}^a, \hat{\mathbf{f}}_p^v, \hat{\mathbf{c}}_i^a, \hat{\mathbf{c}}_i^v\in\mathbb{R}^{1\times D}$, and $D$ denotes the dimension of embeddings.
$[\ ;\ ]$ is the concatenation operator. 
The self-attention operator $\phi^t_a(\cdot)$ is formulated as:
\begin{equation}
    \phi^t_a(\mathbf{x}^t_{a,j}, \mathbf{X}^t_a, \mathbf{X}^t_a) = \mbox{Softmax}(\dfrac{\mathbf{x}^t_{a,j}{\mathbf{X}^t_a}^\top}{\sqrt{D}})\mathbf{X}^t_a
\end{equation}
\begin{equation}
    \phi^t_v(\mathbf{x}^t_{v,j}, \mathbf{X}^t_v, \mathbf{X}^t_v) = \mbox{Softmax}(\dfrac{\mathbf{x}^t_{v,j}{\mathbf{X}^t_v}^\top}{\sqrt{D}})\mathbf{X}^t_v
\end{equation}
Then, to avoid forgetting old class tokens $\{\mathbf{c}^t_i\}_{i=1}^{C}$ at task $t-1$, we apply a Kullback-Leibler (KL) divergence loss $\mbox{KL}(\mathbf{c}^t_i||\mathbf{c}^{t-1}_i)$ on new task $t$ and previous task $t-1$.
Meanwhile, to constrain the independence of each new class token $\{\mathbf{c}^t_i\}_{i=C+1}^{2C}$, we use a fully-connected (FC) layer and add a softmax operator to predict the new class probability: 
$\mathbf{e}_i^t = \mbox{Softmax}(\textsc{FC}(\mathbf{c}_i^t))$. 
Then, a cross-entropy loss $\sum_{i=1}^C\mbox{CE}(\mathbf{h}_i^t, \mathbf{e}_i^t)$ optimizes each new audio-visual category probability, where $\mbox{CE}(\cdot)$ denote cross-entropy loss; $\mathbf{h}_i^t$ is a one-hot encoding with its element as 1 in the target class entry $i$.
Optimizing the KL and cross-entropy loss together will push the learned token embeddings discriminative.

\begin{table*}[t]
	\renewcommand\tabcolsep{6.0pt}
	\centering
	\scalebox{0.8}{
		\begin{tabular}{l|cc|cc|cc}
			\toprule
			\multirow{2}{*}{Method} & \multicolumn{2}{c|}{Audio} & \multicolumn{2}{c|}{Visual} & \multicolumn{2}{c}{Audio-Visual}  \\
			 & Average Acc $\uparrow$(\%) &  Forgetting $\downarrow$(\%) & Average Acc $\uparrow$(\%) &  Forgetting $\downarrow$(\%) & Average Acc $\uparrow$(\%) &  Forgetting $\downarrow$(\%) \\ 		
			\midrule
			EWC~\cite{Kirkpatrick2017overcoming} & 17.32 & 37.28 & 15.19 & 39.75 & 23.27 & 33.65 \\

                LwF~\cite{li2018learning} & 26.37 & 32.56 & 25.21 & 33.16 & 32.16 & 27.13 \\				

                iCaRL~\cite{Rebuffi2017icarl} & 27.25 & 30.75 & 28.32 & 29.37 & 33.26 & 26.53 \\

                IL2M~\cite{Belouadah2019il2m} & 31.65 & 26.72 & 30.29 & 28.63 & 38.13 & 21.25 \\

                BiC~\cite{wu2019large} & 38.23 & 22.38 & 35.72 & 24.86 & 45.23 & 16.38 \\

                PODNet~\cite{Douillard2020podnet} & 42.27 & 19.52 & 40.51 & 21.09 & 47.17 & 14.87 \\

                SS-IL~\cite{Ahn2021ssil} & 47.05 & 15.09 & 45.59 & 16.83 & 53.26 & 13.16 \\

                L2P~\cite{wang2022l2p} & 53.06 & 11.05 & 51.57 & 11.39 & 59.15 & 9.76 \\

                DualPrompt~\cite{wang2022dualprompt} & 57.27 & 8.28 & 56.35 & 8.52 & 63.32 & 6.03 \\

                CIGN (ours) & \textbf{62.53} & \textbf{5.57} & \textbf{60.37} & \textbf{5.98} & \textbf{70.06} & \textbf{2.62} \\ \hline
                Upper-bound & 65.23 & -- & 63.57 & -- & 73.68 & -- \\
			\bottomrule
			\end{tabular}}
   \caption{Quantitative results of audio, visual, and audio-visual continual learning on VGGSound-Instruments dataset.}
   \label{tab: exp_sota_instruments}
\end{table*}

\subsection{Audio-Visual Continual Grouping}\label{sec:avcg}
With the benefit of the above constraints on learning old and new categories, we propose a novel and explicit audio-visual continual grouping module composed of grouping blocks $g^t_a(\cdot), g^t_v(\cdot)$ to take the learned audio-visual incremental category tokens and aggregated features as inputs to generate category-aware incremental embeddings as:
\begin{equation}
\begin{aligned}
    \{\mathbf{g}^t_{a,i}\}_{i=1}^{2C}  &= g^t_a(\{\hat{\mathbf{f}}^t_a, \{\hat{\mathbf{c}}_i^a\}_{i=1}^{2C}),\\
    \{\mathbf{g}^t_{v,i}\}_{i=1}^{2C}  &= g^t_v(\{\hat{\mathbf{f}}^t_{v,p}\}_{p=1}^P, \{\hat{\mathbf{c}}^t_{v,i}\}_{i=1}^{2C})
\end{aligned}
\end{equation}
During the grouping stage, we merge all the audio-visual embeddings with the same category token into an updated class-aware audio-visual feature by computing the global audio similarity vector $\mathbf{A}^t_a\in\mathbb{R}^{1\times 2C}$ and spatial visual similarity matrix $\mathbf{A}^t_v\in\mathbb{R}^{P\times 2C}$ between audio-visual features and $2C$ class tokens at task $t$ via a softmax operation, which is formulated as
\begin{equation}
\begin{aligned}
    & \mathbf{A}^t_{a,i} = \mbox{Softmax}(W^t_{q,a}\hat{\mathbf{f}}^t_a \cdot W^t_{k,a}\hat{\mathbf{c}}^t_{a,i}), \\
    & \mathbf{A}^t_{v,p,i} = \mbox{Softmax}(W^t_{q,v}\hat{\mathbf{f}}^t_{v,p} \cdot W^t_{k,v}\hat{\mathbf{c}}^t_{v,i})
\end{aligned}
\end{equation}
where $W^t_{q,a}, W^t_{k,a}\in\mathbb{R}^{D\times D}$ and $W^t_{q,v}, W^t_{k,v}\in\mathbb{R}^{D\times D}$ denote learnable weights of linear projections for audio-visual features and class tokens at task $t$, respectively.
Using this global audio similarity vector and spatial visual similarity matrix, we calculate the weighted sum of audio-visual features belonged to generate the class-aware embeddings as:
\begin{equation}\label{eq:uni_group}
\begin{aligned}
    \mathbf{g}^t_{a,i} & = g^t_a(\hat{\mathbf{f}}^t_a, \hat{\mathbf{c}}^t_{a,i})=\hat{\mathbf{c}}^t_{a,i} + W^t_{o,a}\dfrac{\mathbf{A}^t_{a,i}W^t_{v,a}\hat{\mathbf{f}}^t_a}{\mathbf{A}^t_{a,i}}, \\
    \mathbf{g}^t_{v,i} & = g^t_v(\{\hat{\mathbf{f}}^t_{v,p}\}_{p=1}^P, \hat{\mathbf{c}}^t_{v,i}) = \hat{\mathbf{c}}^t_{v,i} + W^t_{o,v}\dfrac{\sum_{p=1}^{P}\mathbf{A}^t_{v,p,i}W^t_{v,v}\hat{\mathbf{f}}^t_{p,v}}{\sum_{p=1}^{P}\mathbf{A}^t_{v,p,i}}
\end{aligned}
\end{equation}
where $W^t_{o,a}, W^t_{v,a}\in\mathbb{R}^{D\times D}$ and $W^t_{o,v}, W^t_{v,v}\in\mathbb{R}^{D\times D}$ denote learned weights of linear projections for output and value in terms of audio-visual modalities at task $t$, separately. 
With class-aware audio-visual features $\{\mathbf{g}^t_{a,i}\}_{i=1}^{2C}, \{\mathbf{g}^t_{v,i}\}_{i=1}^{2C}$ as the inputs, we apply an FC layer and sigmoid operator on them to predict the binary probability: $p^t_{a,i} = \mbox{Sigmoid}(\textsc{FC}(\mathbf{g}^t_{a,i})), p^t_{v,i} = \mbox{Sigmoid}(\textsc{FC}(\mathbf{g}^t_{v,i}))$ for $i$th class. 
By applying audio-visual incremental classes $\{y_i^t\}_{i=1}^{2C}$ as the supervision and combining the constraint loss, we formulate a class-constrained grouping loss as:
\begin{equation}
\begin{aligned}
    \mathcal{L}_{\mbox{group}} = & \sum_{i=1}^{C}\mbox{KL}(\mathbf{c}^t_i||\mathbf{c}^{t-1}_i)+ \sum_{i=C+1}^{2C}\mbox{CE}(\mathbf{h}_i^t, \mathbf{e}_i^t) \\
    & + \sum_{i=1}^{2C} \{\mbox{BCE}(y_i^t, p^t_{a,i}) + \mbox{BCE}(y_i^t, p^t_{v,i})\}.
\end{aligned}
\end{equation}
With the help of the introduced class-constrained objective, we generate category-aware audio-visual representations $\{\mathbf{g}^t_{a,i}\}_{i=1}^{2C}, \{\mathbf{g}^t_{v,i}\}_{i=1}^{2C}$ for audio-visual alignment.
Note that global audio-visual features for $2N$ class tokens $\{\mathbf{g}^t_{a,n}\}_{n=1}^{2N}, \{\mathbf{g}^t_{v,n}\}_{n=1}^{2N}$ are chosen from $2C$ categories according to the associated ground-truth class. 
Therefore, the audio-visual similarities between old and new classes are computed by max-pooling audio-visual cosine similarities of class-aware audio-visual features $\{\mathbf{g}^{t-1}_{a,n}\}^{N}_{n=1}, \{\mathbf{g}^{t-1}_{v,n}\}^{N}_{n=1}$ at task $t-1$ and those features from old classes $\{\mathbf{g}^{t}_{a,n}\}^{N}_{n=1}, \{\mathbf{g}^{t}_{v,n}\}^{N}_{n=1}$ and new classes $\{\mathbf{g}^{t}_{a,n}\}^{2N}_{n=N+1}, \{\mathbf{g}^{t}_{v,n}\}^{2N}_{n=N+1}$ at task $t$.
With these category-aware similarities, we formulate the continual audio-visual grouping loss as:
\begin{equation}\label{eq:micl}
\begin{aligned}
    \mathcal{L}_{\mbox{ctl}} = 
    & - \frac{1}{N}\sum_{n=1}^N \log \frac{
    \exp \left( \frac{1}{\tau} \mathtt{sim}(\mathbf{g}^{t-1}_{a,n}, \mathbf{g}^{t}_{a,n}) \right)
    }{
    \sum_{m=N+1}^{2N} \exp \left(  \frac{1}{\tau} \mathtt{sim}(\mathbf{g}^{t-1}_{a,n}, \mathbf{g}^{t}_{a,m})\right)} \\
    & - \frac{1}{N}\sum_{n=1}^N \log \frac{
    \exp \left( \frac{1}{\tau} \mathtt{sim}(\mathbf{g}^{t-1}_{v,n}, \mathbf{g}^{t}_{v,n}) \right)
    }{
    \sum_{m=N+1}^{2N} \exp \left(  \frac{1}{\tau} \mathtt{sim}(\mathbf{g}^{t-1}_{v,n}, \mathbf{g}^{t}_{v,m})\right)} 
\end{aligned}
\end{equation}
The overall objective of our model is simply optimized in an end-to-end manner as:
\begin{equation}
    \mathcal{L} = \mathcal{L}_{\mbox{ctl}} + \mathcal{L}_{\mbox{group}}
\end{equation}
During inference, we follow the prior work~\cite{wang2022l2p,wang2022dualprompt} and use one single model with parameters trained at task $t$ for evaluation, and the product of output logits ($p^t_{a,i},p^t_{v,i}$) from audio and visual modalities are utilized to predict the probability of audio-visual classes, that is $p^t_{av,i} = p^t_{a,i}\cdot p^t_{v,i}$.

\begin{table*}[t]
	\renewcommand\tabcolsep{6.0pt}
	\centering
	\scalebox{0.8}{
		\begin{tabular}{l|cc|cc|cc}
			\toprule
			\multirow{2}{*}{Method} & \multicolumn{2}{c|}{Audio} & \multicolumn{2}{c|}{Visual} & \multicolumn{2}{c}{Audio-Visual}  \\
			 & Average Acc $\uparrow$(\%) &  Forgetting $\downarrow$(\%) & Average Acc $\uparrow$(\%) &  Forgetting $\downarrow$(\%) & Average Acc $\uparrow$(\%) &  Forgetting $\downarrow$(\%) \\ 		
			\midrule
			EWC~\cite{Kirkpatrick2017overcoming} & 15.17 & 43.05 & 18.53 & 41.25 & 21.52 & 39.16 \\

                LwF~\cite{li2018learning} & 24.25 & 33.83 & 26.78 & 32.56 & 29.16 & 28.25 \\				

                iCaRL~\cite{Rebuffi2017icarl} & 23.55 & 35.32 & 27.31 & 33.07 & 28.37 & 31.02 \\

                IL2M~\cite{Belouadah2019il2m} & 26.32 & 27.56 & 30.25 & 25.83 & 32.65 & 23.19 \\

                BiC~\cite{wu2019large} & 32.31 & 22.13 & 35.09 & 19.35 & 38.56 & 17.02 \\

                PODNet~\cite{Douillard2020podnet} & 33.56 & 21.23 & 37.16 & 17.56 & 40.03 & 15.51 \\

                SS-IL~\cite{Ahn2021ssil} & 36.21 & 18.75 & 40.57 & 16.82 & 43.16 & 13.89 \\				

                L2P~\cite{wang2022l2p} & 38.05 & 16.13 & 42.06 & 13.81 & 47.82 & 11.05 \\

                DualPrompt~\cite{wang2022dualprompt} & 42.25 & 13.58 & 46.28 & 10.03 & 49.72 & 9.63 \\

                CIGN (ours) & \textbf{45.83} & \textbf{10.21} & \textbf{49.52} & \textbf{8.83} & \textbf{55.26} & \textbf{6.52} \\	\hline		

                Upper-bound & 50.06 & -- & 53.32 & -- & 60.57 & -- \\
			\bottomrule
			\end{tabular}}
   \caption{Quantitative results of audio, visual, and audio-visual continual learning on VGGSound-100 dataset.}
   \label{tab: exp_sota_vgg100}
\end{table*}

\section{Experiments}

\subsection{Experimental Setup}

\noindent\textbf{Datasets.}
VGGSound-Instruments~\cite{hu2022mix} includes 32k video clips of 10s lengths from 36 musical instrument classes, a subset of VGG-Sound~\cite{chen2020vggsound}, and each video only has one single instrument class annotation. 
Beyond this instrumental benchmark, we filter 97k video clips of 10s lengths from the original VGG-Sound~\cite{chen2020vggsound}, denoted as VGGSound-100, and consists of 100 categories, such as nature, animals, vehicles, human speech, dancing, playing instruments, etc.
For large-scale incremental learning, we use the full VGG-Sound Source~\cite{chen2021localizing} with 150k video clips with 220 categories in the original VGG-Sound~\cite{chen2020vggsound}.
Each dataset is split into train/val/test  sets with respective ratios of 0.8/0.1/0.1.

\noindent\textbf{Evaluation Metrics.}
Following the prior work~\cite{Ahn2021ssil,wang2022l2p,wang2022dualprompt}, we use the class-incremental setting of $T=4$ sequential classification tasks with equal sizes of categories, where each task has a separate test set.
With the common metrics in previous methods~\cite{wang2022l2p,wang2022dualprompt}, we apply Average accuracy and Forgetting for comprehensive evaluation.
Higher Average accuracy is better, and lower Forgetting is better.

\noindent\textbf{Implementation.}
For input frames, we resize the resolution to $224 \times 224$. 
For input audio, the log spectrograms are generated from $3s$ of audio with a sample rate of $22050$Hz. 
Following the prior work~\cite{mo2022EZVSL,mo2022SLAVC}, we apply STFT to extract an input tensor of size $257 \times 300$ ($257$ frequency bands over $300$ timesteps) by using 50ms windows at a hop size of 25ms. 
We follow previous work~\cite{hu2019deep,qian2020multiple,chen2021localizing,mo2022EZVSL,mo2022SLAVC}, we apply the lightweight ResNet18~\cite{he2016resnet} as the audio and visual encoder, and initialize them using weights pre-trained on 2M Flickr videos~\cite{aytar2016soundnet} using the state-of-the-art self-supervised source localization approach~\cite{mo2022EZVSL}.
The input dimension $D$ of embeddings is $512$, and the total number $P$ of patches for the $7\times 7$ spatial map is 49.
The depth of self-attention transformers $\phi^a(\cdot),\phi^v(\cdot)$ is 3 as default.
We train the model for 100 epochs with a batch size of 128.
The Adam optimizer~\cite{kingma2014adam} is used with a learning rate of $1e-4$.
Following the prior work~\cite{wang2022l2p}, we randomly use $50$ audio-visual pairs per category from old tasks for the buffer size.

\begin{table*}[t]
	\renewcommand\tabcolsep{6.0pt}
	\centering
	\scalebox{0.8}{
		\begin{tabular}{l|cc|cc|cc}
			\toprule
			\multirow{2}{*}{Method} & \multicolumn{2}{c|}{Audio} & \multicolumn{2}{c|}{Visual} & \multicolumn{2}{c}{Audio-Visual}  \\
			 & Average Acc $\uparrow$(\%) &  Forgetting $\downarrow$(\%) & Average Acc $\uparrow$(\%) &  Forgetting $\downarrow$(\%) & Average Acc $\uparrow$(\%) &  Forgetting $\downarrow$(\%) \\ 		
			\midrule
			EWC~\cite{Kirkpatrick2017overcoming} & 9.28 & 59.03 & 13.51 & 52.16 & 16.36 & 46.72 \\
                LwF~\cite{li2018learning} & 17.67 & 48.36 & 20.32 & 45.36 & 25.72 & 38.65 \\
                iCaRL~\cite{Rebuffi2017icarl} & 18.56 & 46.28 & 21.03 & 44.52 & 23.56 & 42.35 \\

                IL2M~\cite{Belouadah2019il2m} & 23.07 & 41.76 & 26.51 & 36.32 & 30.25 & 31.86 \\

                BiC~\cite{wu2019large} & 25.68 & 37.83 & 29.26 & 31.33 & 33.19 & 23.56 \\

                PODNet~\cite{Douillard2020podnet} & 28.23 & 32.57 & 31.06 & 29.71 & 36.51 & 18.62 \\

                SS-IL~\cite{Ahn2021ssil} & 31.56 & 28.16 & 35.52 & 26.62 & 40.21 & 15.23 \\

                L2P~\cite{wang2022l2p} & 34.21 & 25.19 & 36.21 & 20.07 & 42.15 & 13.52 \\

                DualPrompt~\cite{wang2022dualprompt} & 37.52 & 22.63 & 40.86 & 17.27 & 43.61 & 11.56 \\

                CIGN (ours) & \textbf{39.81} & \textbf{15.25} & \textbf{42.26} & \textbf{13.01} & \textbf{46.58} & \textbf{8.67} \\ \hline
                Upper-bound & 45.89 & -- & 51.56 & -- & 53.75 & -- \\
			\bottomrule
			\end{tabular}}
   \caption{Quantitative results of audio, visual, and audio-visual continual learning on VGG-Sound Sources dataset.}
   \label{tab: exp_sota_vggsound}
\end{table*}

\begin{table*}[t]
	\renewcommand\tabcolsep{6.0pt}
    \renewcommand{\arraystretch}{1.1}
	\centering
	\scalebox{0.85}{
		\begin{tabular}{cccccccc}
			\toprule
			\multirow{2}{*}{AVCTD} & \multirow{2}{*}{AVCG} & \multicolumn{2}{c}{Audio} & \multicolumn{2}{c}{Visual} & \multicolumn{2}{c}{Audio-Visual} \\
			& & Average Acc $\uparrow$(\%) &  Forgetting $\downarrow$(\%) & Average Acc $\uparrow$(\%) &  Forgetting $\downarrow$(\%) & Average Acc $\uparrow$(\%) &  Forgetting $\downarrow$(\%) \\ 	
			\midrule
			\xmark & \xmark & 36.72 & 19.23 & 39.35 & 17.23& 	45.05 & 	15.26 \\

			\cmark & \xmark & 42.59 & 13.21 & 47.12 & 10.72& 	51.02 & 	9.75 \\

			\xmark & \cmark & 40.16 & 15.03 & 43.56 & 13.69 & 	48.72 & 	11.57 \\

			\cmark & \cmark & \textbf{45.83} & \textbf{10.21} & \textbf{49.52} & \textbf{8.83} & \textbf{55.26} & \textbf{6.52} \\
			\bottomrule
			\end{tabular}}
   \caption{Ablation studies on Audio-Visual Class Tokens Distillation (AVCTD) and Audio-Visual Continual Grouping (AVCG). }
	\label{tab: exp_ablation}
			\vspace{-0.5em}
\end{table*}

\subsection{Comparison to Prior Work}\label{sec:exp}

In this work, we propose a novel and effective framework for continual audio-visual learning.
In order to demonstrate the effectiveness of the proposed CIGN, we comprehensively compare it to previous continual learning baselines:
1) EWC~\cite{Kirkpatrick2017overcoming} (2017'PNAS): 
a vanilla baseline that addressed catastrophic forgetting in neural networks;
2) LwF~\cite{li2018learning} (2018'TPAMI): a regularization-based approach for training new task data and preserving capabilities of old tasks;
3) iCaRL~\cite{Rebuffi2017icarl} (2017'CVPR):  a class-incremental work that continuously used a sequential data stream for new classes;
4) IL2M~\cite{Belouadah2019il2m} (2019'ICCV): a dual memory network that combined a bounded memory of the past images and a second memory for past class statistics;
5) BiC~\cite{wu2019large} (2019'CVPR): a bias correction method that tackled the imbalance between old and new classes;
6) PODNet~\cite{Douillard2020podnet} (2020'ECCV):
a spatial-based distillation framework with proxy vectors for each category;
7) SSIL~\cite{Ahn2021ssil} (2021'ICCV): a task-wise knowledge distillation network based on the separated softmax output layer;
8) L2P~\cite{wang2022l2p} (2022'CVPR): a strong prompts-based baseline optimizing the prompt pool memory as instructions for sequential tasks;
9) DualPrompt~\cite{wang2022dualprompt} (2022'ECCV):
a recent strong baseline with General-Prompt and Expert-Prompt for task-invariant and task-specific knowledge;
10) Upper-bound: a usually supervised baseline training on the data of all tasks.

For the VGGSound-Instruments dataset, we report the quantitative comparison results in Table~\ref{tab: exp_sota_instruments}.
As can be seen, we achieve the best results regarding all metrics for three class-incremental settings (audio, visual, and audio-visual) compared to previous continual learning approaches.
In particular, the proposed CIGN superiorly outperforms DualPrompt~\cite{wang2022dualprompt}, the current state-of-the-art continual learning baseline, by 5.26 Average Acc@Audio \& 2.71 Forgetting@Audio, 4.02 Average Acc@Visual \& 2.54 Forgetting@Visual, and 6.74 Average Acc@Audio-Visual \& 3.41 Forgetting@Audio-Visual on three settings.
Furthermore, we achieve significant performance gains compared to L2P~\cite{wang2022l2p}, the current state-of-the-art rehearsal-based baseline, which indicates the importance of extracting category-aware semantics from incremental audio-visual inputs as guidance for continual audio-visual learning.
Meanwhile, the gap between our CIGN and the oracle performance of upper-bound using all data for training is the lowest compared to other baselines.
These significant improvements demonstrate the superiority of our approach in audio-visual class-incremental learning.

In addition, significant gains in VGGSound-100 and VGG-Sound Sources benchmarks can be observed in Table~\ref{tab: exp_sota_vgg100} and Table~\ref{tab: exp_sota_vggsound}.
Compared to L2P~\cite{wang2022l2p}, the current state-of-the-art rehearsal-based method, we achieve the results gains of 7.78 Average Acc@Audio, 7.46 Average Acc@Visual, and 7.44 Average Acc@Audio-Visual on VGGSound-100 dataset.
Moreover, when evaluated on the challenging VGG-Sound Sources benchmark, the proposed method still outperforms L2P~\cite{wang2022l2p} by 5.60 Average Acc@Audio, 6.05 Average Acc@Visual, and 4.43 Average Acc@Audio-Visual.
We also achieve highly better results against SS-IL~\cite{Ahn2021ssil}, the task-wise knowledge distillation network based on separated softmax.
These results demonstrate the effectiveness of our approach in learning disentangled semantics from incremental audio and images for continual audio-visual classification.

\subsection{Experimental Analysis}

In this section, we performed ablation studies to demonstrate the benefit of introducing the Audio-Visual Class Tokens Distillation and Audio-Visual Continual Grouping modules. 
We also conducted extensive experiments to explore a flexible number of incremental tasks, and learned disentangled category-aware audio-visual representations.

\vspace{2mm}
\noindent\textbf{Audio-Visual Class Tokens Distillation \& Audio-Visual Continual Grouping.}
In order to demonstrate the effectiveness of the introduced audio-visual class tokens distillation (AVCTD) and audio-visual continual grouping (AVCG), we ablate the necessity of each module and report the quantitative results on VGGSound-100 dataset in Table~\ref{tab: exp_ablation}.
As can be observed, adding AVCTD to the vanilla baseline highly increases the results of Average Acc (by 5.87@Audio, 7.77@Visual, and 5.97@Audio-Visual) and decreases the performance of forgetting (by 6.02@Audio, 6.51@Visual, and 5.51@Audio-Visual), which validates the benefit of category tokens distillation in extracting disentangled high-level semantics for class-incremental learning. 
Meanwhile, introducing only AVCG in the baseline increases the class-incremental learning performance regarding all metrics.
More importantly, incorporating AVCTD and AVCG into the baseline significantly raises the performance of Average Acc by 9.11@Audio, 10.17@Visual, and 10.21@Audio-Visual, and reduces the results of Forgetting by 9.02@Audio, 8.40@Visual, and 8.74@Audio-Visual. 
These improving results validate the importance of audio-visual class tokens distillation and audio-visual continual grouping in extracting category-aware semantics from class-incremental audio and image for continual audio-visual learning.

\begin{figure*}[t]
\centering
\includegraphics[width=0.98\linewidth]{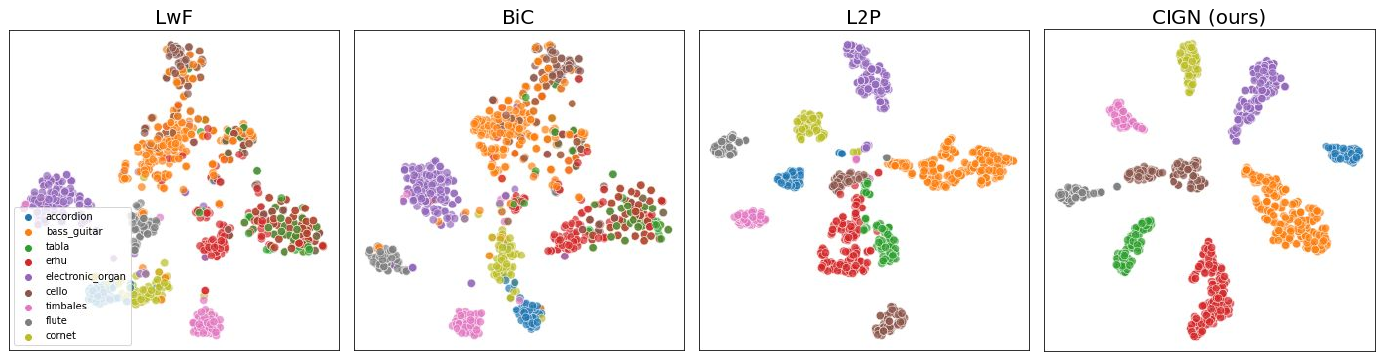}
\caption{Qualitative comparisons of representations learned by LwF, BiC, L2P, and the proposed CIGN. 
Note that each spot denotes features extracted from one sample, and each color refers to one audio-visual category, such as ``cello" in brown and ``tabla" in green.}
\label{fig: exp_vis_feat}
\vspace{-0.5em}
\end{figure*}
\vspace{2mm}
\noindent\textbf{Generalizing to Flexible Number of Incremental Tasks.}
In order to validate the generalizability of the proposed CIGN to a flexible number of incremental tasks, we transfer the model to continual training on 10 tasks with every 10 categories on the VGGSound-100 benchmark.
We still achieve competitive results of Average Acc (40.53@Audio, 43.29@Visual, 48.16@Audio-Visual) and Forgetting (13.87@Audio, 12.16@Visual, 9.75@Audio-Visual) on the challenging VGGSound-100 dataset.
These results indicate that our CIGN can support a flexible number of incremental classification tasks, which further demonstrates the effectiveness of our class-incremental grouping network for continual audio-visual learning.

\vspace{2mm}
\noindent\textbf{Learned Category-aware Audio-Visual Representations.}
Learning disentangled audio-visual representations with category-aware semantics is critical for classifying audio-visual pairs from incremental classes.
To better evaluate the quality of learned category-aware features, we visualize the learned visual and audio representations of 9 categories in the first task after finishing 4 incremental tasks on VGGSound-Instruments by t-SNE~\cite{laurens2008visualizing}, as shown in Figure~\ref{fig: exp_vis_feat}.
It should be noted that each color denotes one class of the audio-visual pair, such as ``cello'' in brown and ``tabla'' in green.
As can be seen in the last column, audio-visual embeddings extracted by the proposed CIGN are both intra-class compact and inter-class separable. 
In contrast to our disentangled representations in the audio-visual semantic space, mixtures of multiple audio-visual categories still exist among features learned by LwF~\cite{li2018learning} and BiC~\cite{wu2019large}.
With the help of text prompts for images, L2P~\cite{wang2022dualprompt} can extract clustered audio-visual incremental representations on some classes, such as ``base\_guitar'' in orange. 
However, some categories are mixed while others are very close to each other as they do not incorporate the explicit audio-visual continual grouping mechanism in our CIGN.
These meaningful visualization results further showcase the superiority of our CIGN in extracting compact audio-visual incremental representations with class-aware semantics for continual audio-visual learning.

\subsection{Limitation}

Although the proposed CIGN achieves superior results on both single-modality and cross-modal class-incremental learning, the performance gains of our method on the VGGSound-Instruments benchmark with a limited size of buffers are not substantial.
One possible cause is that our model overfits during the training time, and the solution is to incorporate dropout and momentum encoders together for continual audio-visual classification.
Meanwhile, we observe that if we transfer our model to open-set audio-visual classification without additional training, it would be hard to predict unseen categories as we need to pre-define a set of categories during training and do not learn unseen class tokens to guide the classification.
Future work could add enough learnable category tokens for rehearsal-free continual learning in new categories.

\section{Conclusion}

In this work, we present CIGN, a novel class-incremental grouping network that can directly learn category-wise semantic features to achieve continual audio-visual learning.
We leverage learnable audio-visual class tokens and audio-visual grouping to aggregate class-aware features continually.
Furthermore, we introduce class tokens distillation and continual grouping to alleviate forgetting parameters learned from previous tasks for capturing discriminative audio-visual categories.
Experimental results on VGGSound-Instruments, VGGSound-100, and VGG-Sound Sources benchmarks comprehensively demonstrate the state-of-the-art superiority against previous regularization- and rehearsal-based class-incremental learning baselines on continual audio-visual settings. 
Meanwhile, qualitative visualizations of incremental audio-visual embeddings vividly showcase the effectiveness of our CIGN in aggregating class-aware features to avoid cross-modal catastrophic forgetting. 
Extensive ablation studies also validate the importance of class tokens distillation and continual grouping in learning compact representations for continual audio-visual learning.

\vspace{2mm}
\noindent\textbf{Broader Impact.}
The proposed method predicts class-incremental audio-visual pairs learning from web videos, which could cause the model to learn internal biases in the data. 
For instance, the model might fail to predict certain rare but crucial audio-visual classes. 
These issues should be incrementally resolved for real-world applications.

\vspace{2mm}
\noindent
\textbf{Acknowledgments.} We would like to thank the anonymous reviewers for their constructive comments. This work was supported in part by gifts from Cisco Systems and Adobe. The article solely reflects the opinions and conclusions of its authors but not the funding agents.

{\small
\bibliographystyle{ieee_fullname}
\bibliography{reference}
}

\newpage

\appendix
\section*{Appendix}

In this supplementary material, we provide the significant differences between our CIGN and the recent grouping work, GroupViT~\cite{xu2022groupvit}, and more experiments on the depth of transformer layers and grouping strategies. 
In addition, we validate the effectiveness of learnable audio-visual class tokens in learning disentangled class-incremental audio-visual representations for continual audio-visual learning and report quantitative comparison results of various buffer sizes.

\begin{table*}[t]
	\renewcommand\tabcolsep{6.0pt}
    \renewcommand{\arraystretch}{1.1}
	\centering
	\scalebox{0.85}{
		\begin{tabular}{cccccccc}
			\toprule
			\multirow{2}{*}{Depth} & \multirow{2}{*}{AVCG} & \multicolumn{2}{c}{Audio} & \multicolumn{2}{c}{Visual} & \multicolumn{2}{c}{Audio-Visual} \\
			& & Average Acc $\uparrow$(\%) &  Forgetting $\downarrow$(\%) & Average Acc $\uparrow$(\%) &  Forgetting $\downarrow$(\%) & Average Acc $\uparrow$(\%) &  Forgetting $\downarrow$(\%) \\ 	
			\midrule
			1 & Softmax & 42.25 & 13.09 & 47.26 & 10.63& 	51.25  & 	9.31 \\
			3 & Softmax & \textbf{45.83} & \textbf{10.21} & \textbf{49.52} & \textbf{8.83} & \textbf{55.26} & \textbf{6.52} \\
   6 & Softmax & 44.62 & 11.52 & 48.63 & 9.55	& 53.21	  & 7.58 \\ 
   12 & Softmax & 43.91 & 12.36 & 48.01 & 10.12& 	52.53  & 	8.72 \\ 
3& Hard-softmax & 40.59 & 15.16 & 43.72 & 13.52& 	48.95  & 	11.36 \\ 
			\bottomrule
			\end{tabular}}
   \caption{Exploration studies on the depth of self-attention transformer layers and continual grouping strategies in Audio-Visual Continual Grouping (AVCG) module. }
	\label{tab: ab_depth}
			\vspace{-0.5em}
\end{table*}

\section{Significant Difference from GroupViT and CIGN}

Compared to GroupViT~\cite{xu2022groupvit} on image segmentation, our CIGN has three significant recognizable characteristics to address continual cross-modal learning from incremental categories of audio-visual pairs, which are highlighted as follows:

1) \textbf{Incremental-Constraint on Audio-Visual Category Tokens.}
The major difference is that we have learned disentangled audio-visual class tokens for each audio category, \textit{e.g.}, 100 audio-visual category tokens for 100 categories in the VGGSound-100 benchmark.
During training, each audio-visual class token does not learn semantic overlapping information among each other, where we apply the cross-entropy loss $\sum_{i=1}^C\mbox{CE}(\mathbf{h}_i^t, \mathbf{e}_i^t)$ on each category probability $\mathbf{e}_i^t$ with the disentangled constraint $\mathbf{h}_i^t$ at current task $t$.
Meanwhile, we apply a Kullback-Leibler (KL) divergence loss $\mbox{KL}(\mathbf{c}^t_i||\mathbf{c}^{t-1}_i)$ to eliminate forgetting old class tokens $\{\mathbf{c}^t_i\}_{i=1}^{C}$ at task $t-1$.
However, the number of group tokens used in GroupViT is a hyper-parameter, and they must tune it carefully across each grouping stage.

2) \textbf{Audio-Visual Continual Grouping.}
We propose the audio-visual continual grouping module for extracting individual semantics with class-aware information from incremental audio-visual pairs. 
However, GroupViT utilized the grouping mechanism on only patches of images without explicit category-aware tokens involved. 
Therefore, GroupViT can not be directly transferred to incremental audio-visual samples for solving the new continual audio-visual learning problem. 
Moreover, they used multiple grouping stages during training, and the number of grouping stages is a hyper-parameter. 
In our grouping module, only one audio-visual incremental grouping stage with disentangled and incremental audio-visual class tokens is enough to capture disentangled audio-visual representations in the multi-modal incremental semantic space.

3) \textbf{Incremental Audio-Visual Class as Weak Supervision.}
We leverage the incremental audio-visual category at the current task as the weak supervision to address continual audio-visual learning problem from class-incremental audio-visual samples, while GroupViT used a trivial contrastive loss to match the global visual representations with pre-trained text embeddings. 
In this case, GroupViT required a large batch size for self-supervised training on large-scale language-visual pairs. 
In contrast, we do not need unsupervised learning on the large-scale simulated audio-visual data with extensive training costs.

\section{Depth of Transformer Layers and Continual Grouping Strategies}

The depth of transformer layers and continual grouping strategies used in the proposed AVCG affect the extracted and grouped representations for continual audio-visual learning from incremental cross-modal pairs (\textit{i.e.}, image and audio).
To explore such effects more comprehensively, we varied the depth of transformer layers from $\{1, 3, 6, 12\}$ and ablated the continual grouping strategy using Softmax and Hard-Softmax.
During training, to make Hard-Softmax differentiable, we applied the Gumbel-Softmax~\cite{eric2017categorical,chris2017the} as the alternative.
We report the comparison results of continual audio-visual performance on the VGGSound-100 benchmark in Table~\ref{tab: ab_depth}.
When the depth of transformer layers is 3 and using Softmax in AVCG, we achieve the best class-incremental learning performance regarding all metrics. 
With increased depth from 1 to 3, the proposed CIGN consistently increases performance as better disentangled audio-visual representations are extracted from encoder features of the class-incremental audio-visual samples. 
Nevertheless, increasing the depth from 3 to 12 will not continually improve the class-incremental result since three transformer layers might be enough to extract the learned class-aware representations for audio-visual continual grouping with only one grouping stage.
Furthermore, replacing Softmax with Hard-Softmax significantly deteriorates the performance of all metrics, which indicates the importance of the proposed AVCG in extracting disentangled audio-visual representations with class-incremental 
 category-aware semantics from the audio-visual pairs.

\subsection{Quantitative Validation on Audio-Visual Category Tokens}

Learnable audio-visual incremental category tokens are essential to aggregate audio-visual representations with category-aware semantics from incremental audio-visual samples.
We calculate the Precision, Recall, and F1 scores of audio-visual classification using these representations across training iterations to validate the rationality of learned audio-visual category token embeddings. 
All these metrics are observed to rise to 1, which indicates that each audio-visual category token learned disentangled information with incremental category-aware semantics. 
These quantitative results further demonstrate the effectiveness of audio-visual category tokens distillation in the continual audio-visual grouping for extracting disentangled audio-visual representations from class-incremental audio-visual samples for continual audio-visual learning.

\begin{figure}
\centering
\includegraphics[width=\linewidth]{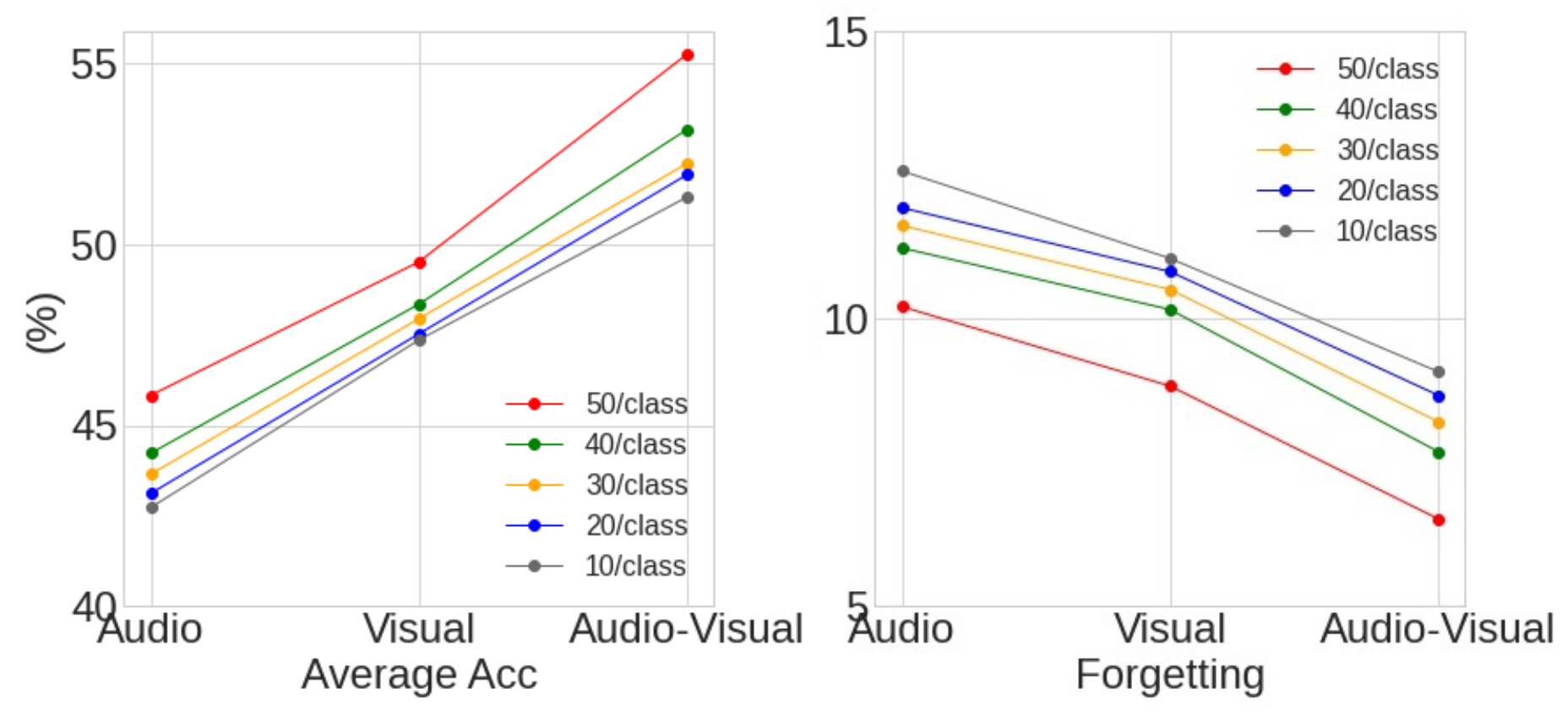}
\caption{Impact of buffer size on the performance of Average Acc (Left) and Forgetting (Right) for continual audio-visual learning.}
\label{fig: ab_buffer}
\vspace{-0.5em}
\end{figure}

\section{Quantitative Comparison on Buffer Size}

To quantitatively demonstrate the effectiveness of buffer size in continual audio-visual learning, we varied the buffer size per class from $\{10, 20, 30, 40, 50\}$, and report the comparison results in Figure~\ref{fig: ab_buffer}.
As can be seen, the proposed CIGN achieves the best performance of average accuracy and forgetting when we use 50 audio-visual samples for each incremental category. 
These results demonstrate the importance of caching samples from previous classes for continual audio-visual learning.

\end{document}